%% file: main.tex
\documentclass[final]{cvpr}

\usepackage{times}
\usepackage{epsfig}
\usepackage{graphicx}
\usepackage{amsmath}
\usepackage{amssymb}

\usepackage{algorithm}
\usepackage{algorithmic}

\usepackage{graphicx} % more modern
\usepackage{caption}
\usepackage{subcaption}
\usepackage{wrapfig}

\usepackage{amsfonts}
\usepackage{multirow}
\usepackage{booktabs}
\usepackage[dvipsnames]{xcolor}
\usepackage{xpatch}
\usepackage{xcolor,colortbl}
\usepackage{array, boldline, makecell}
\usepackage{arydshln}
\usepackage{soul}
\usepackage{stackengine}

\usepackage[pagebackref=true,breaklinks=true,colorlinks,bookmarks=false]{hyperref}

\newcommand\cls[1]{``#1''} % Topic Sentence

\newcommand{\TP}{\mathit{TP}}
\newcommand{\FN}{\mathit{FN}}
\newcommand{\FP}{\mathit{FP}}

\newcommand{\Lcal}{\mathcal{L}}

\newcommand{\thing}{\textit{thing}\xspace}
\newcommand{\stuff}{\textit{stuff}\xspace}
\newcommand{\void}{\textit{void}\xspace}

\newcommand{\unknown}{\textit{unknown}\xspace}
\newcommand{\known}{\textit{known}\xspace}
\newcommand{\seen}{\textit{seen}\xspace}
\newcommand{\unseen}{\textit{unseen}\xspace}

\newcommand{\val}{\textit{val}\xspace}
\newcommand{\train}{\textit{train}\xspace}

\begin{document}

%%%%%%%%% TITLE
\title{Exemplar-Based Open-Set Panoptic Segmentation Network}

\author{Jaedong Hwang$^1$\thanks{This work was done during an internship at Adobe Research.} \qquad Seoung Wug Oh$^2$  \qquad Joon-Young Lee$^2$ \qquad Bohyung Han$^1$\\
$^1$ECE and ASRI, Seoul National University \\
$^2$Adobe Research\\
{\tt\small $^1$\{jd730, bhhan\}@snu.ac.kr, $^2$\{seoh, jolee\}@adobe.com}
}

\maketitle

%%%%%%%%% ABSTRACT
\begin{abstract}
   \input{sections/0_abstract}
\end{abstract}

\input{sections/1_introduction}
\input{sections/2_related}
\input{sections/3_ops}
\input{sections/4_method}
\input{sections/5_exp}
\input{sections/6_discussion}
\input{sections/7_conclusion}

\subsection*{Acknowledgements}
\vspace{-0.2cm}
%\paragraph{Acknowledgement}
\noindent This work was partly supported by Institute for Information \& Communications Technology Promotion (IITP) grant funded by the Korea government (MSIT) [2017-0-01779, 2017-0-01780].

{\small
\bibliographystyle{ieee_fullname}
\bibliography{egbib}
}
\clearpage
\input{sections/8_supp}

\end{document}

%% file: sections/0_abstract.tex
%!TEX root = ./../1361.tex

We extend panoptic segmentation to the open-world and introduce an open-set panoptic segmentation~(OPS) task.
This task requires performing panoptic segmentation for not only \known classes but also \unknown ones that have not been acknowledged during training.
We investigate the practical challenges of the task and construct a benchmark on top of an existing dataset, COCO.
In addition, we propose a novel exemplar-based open-set panoptic segmentation network~(EOPSN) inspired by exemplar theory.
Our approach identifies a new class based on exemplars, which are identified by clustering and employed as pseudo-ground-truths.
The size of each class increases by mining new exemplars based on the similarities to the existing ones associated with the class.
We evaluate EOPSN on the proposed benchmark and demonstrate the effectiveness of our proposals.
The primary goal of our work is to draw the attention of the community to the recognition in the open-world scenarios. 
The implementation of our algorithm is available on the project webpage\footnote{\color{magenta}\url{https://cv.snu.ac.kr/research/EOPSN}}.

%% file: sections/1_introduction.tex
%!TEX root = ./../main.tex

\begin{figure}
\captionsetup[subfigure]{aboveskip=0mm,belowskip=.4mm}
\begin{subfigure}{0.495\linewidth}
 \includegraphics[width=1.0\linewidth]{assets/concept/image.jpeg}
 \subcaption*{Input image}\label{fig:image}
\end{subfigure}
\begin{subfigure}{0.495\linewidth}
 \includegraphics[width=1.0\linewidth]{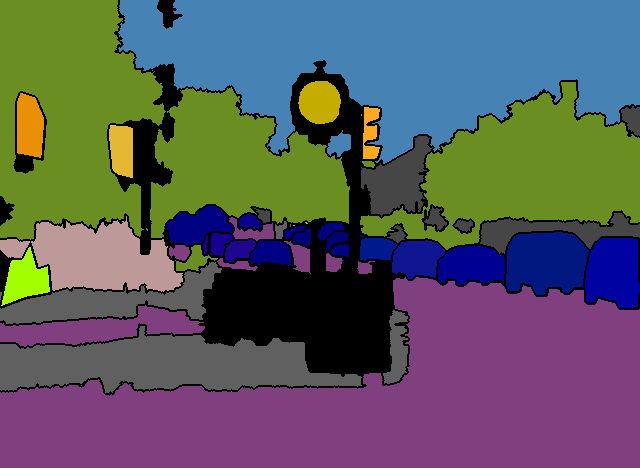}
 \subcaption*{Ground-truth for closed-set}\label{fig:gt}
\end{subfigure}\\
\begin{subfigure}{0.495\linewidth}
 \includegraphics[width=1.0\linewidth]{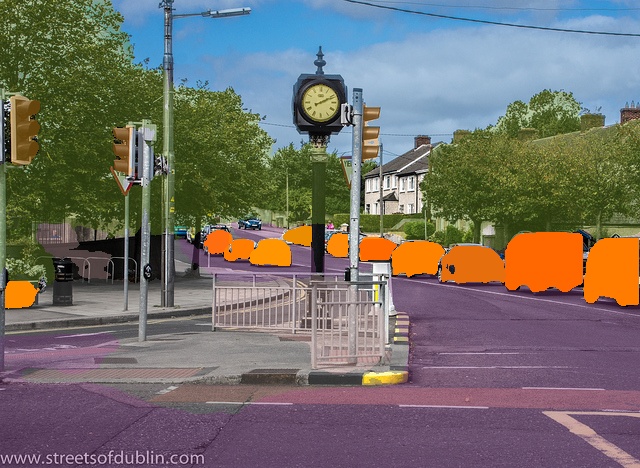}
 \subcaption*{Open-set prediction}\label{fig:pred}
\end{subfigure}
\begin{subfigure}{0.495\linewidth}
 \includegraphics[width=1.0\linewidth]{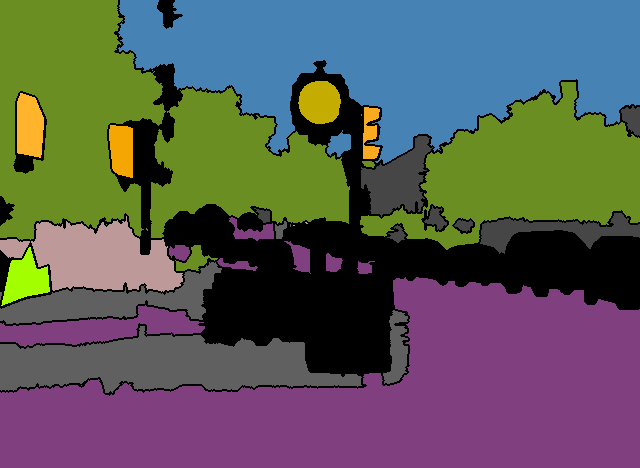}
 \subcaption*{Ground-truth for open-set}\label{fig:gt_open}
\end{subfigure}
\caption{
For a given image, we show the \emph{ground-truth} for closed-set and open-set panoptic segmentation, where unlabeled regions are in black.
In open-set panoptic segmentation, a model needs to find the \unknown class instances, which are not annotated in training data.
We assume that \cls{car} is an \unknown class in the open-set visualization.
\vspace{-0.2cm}
}
\label{fig:concept}
\end{figure}

\section{Introduction}\label{sec:intro}

For deeper visual understanding, researchers have constructed large-scale image-based benchmarks~\cite{cordts2016cityscapes,krishna2017visual,lin2014microsoft,neuhold2017mapillary,olga2015image} and have studied various tasks such as image classification~\cite{he2016deep,szegedy2015going}, object detection~\cite{girshick2015fast,lin2017feature,redmon2017yolo9000,ren2015faster}, semantic segmentation~\cite{chen2018deeplab,long2015fully,noh2015learning}, instance segmentation~\cite{bolya2019yolact,he2017mask}, and many others.
Recently, researchers are getting more interested in finding the location and shape of instances and estimating semantic labels from natural scenes since they are critical for high-level understanding of visual content.

Since Kirillov~\etal~\cite{kirillov2019panoptic} recently formulated a panoptic segmentation task, which is a combination of instance segmentation and semantic segmentation, a number of studies~\cite{cheng2020panoptic,fu2019imp,kirillov2019panopticfpn,lazarow2020learning,li2019attention,liu2019end,porzi2019seamless,xiong2019upsnet,wang2020axial} have been proposed and achieved substantial performance improvements.
However, one drawback of the panoptic segmentation task is the excessive cost of dataset construction.
The annotation of pixel-level panoptic segmentation labels requires significant human efforts, and is even harder than semantic segmentation.
It is challenging to create a large-scale dataset containing such comprehensive annotations.
There are two potential ways to tackle this problem.
The first approach is to develop a weakly supervised panoptic segmentation model~\cite{li2018weakly} that is free from pixel-level annotations.
The other is to build a generalized model that identifies segments of not only trained classes but also unseen ones.
The second approach is closely related to open-set recognition~\cite{scheirer2012toward} that admits the existence of \unknown class in testing.

In this paper, we extend panoptic segmentation to the open-world.
Unlike the closed-set counterpart, the new task has to deal with examples in \unknown classes that are not supervised and acknowledged during training.
Therefore, the goal of this task is to discover panoptic segmentation labels for the examples in both \known and \unknown classes.
We call this new task \emph{open-set panoptic segmentation} (OPS).
We argue that the task is an important milestone for advancing comprehensive visual understanding and it also makes substantial impacts on a wide range of practical applications such as dataset construction, scene analysis, and image editing.
Figure~\ref{fig:concept}  illustrates an example of OPS.

We address the following critical challenges in the journey to OPS and make the task tractable using reasonable assumptions\footnote{Section~\ref{subsec:challenges} discusses the details.}.
The first challenge comes from the definition of \emph{object}.
Since a certain object may consist of many components that belong to another objects, a coarse-grained object class can be split into multiple sub-categories.
For example, a \cls{bus} consists of various parts such as windows, side mirrors, wheels, etc. 
On the other hand, it is difficult to handle the \unknown categories that belong to stuff, \eg, \cls{grass} and \cls{water}, since they are often defined as a region and their individual instances are often ill-defined.
Therefore, an exhaustive panoptic segmentation for all kinds of objects and backgrounds is almost impossible. 
The second issue is that, unlike the open-set image classification that discriminates images in \unknown classes, the open-set panoptic segmentation requires finding \unknown class objects from a scene even when the \unknown class instances, unfortunately, are labeled as background.

We propose an exemplar-based open-set panoptic segmentation network (EOPSN) as a strong baseline model for this new task.
On top of Panoptic FPN~\cite{kirillov2019panopticfpn}, one of the  popular closed-set panoptic segmentation approaches, we integrate an extra component that searches for \unknown class labels inspired by the exemplar theory~\cite{medin1978context,nosofsky1986attention} in psychology.
We first find a new \unknown class and its exemplars by clustering object features and then discover more  based on the similarity to existing exemplars during training.
Note that the proposed method is generic and applicable to any top-down panoptic segmentation method.
Experimental results support the effectiveness of the proposed model; EOPSN outperforms the simple baselines, which are variants of~\cite{kirillov2019panopticfpn}.
The contribution of this paper is three-fold as follows:
\begin{itemize}
    \item {We define the open-set panoptic segmentation (OPS) task and make it feasible using reasonable assumptions through in-depth analysis of its inherent challenges.}
    \item We construct a brand-new OPS benchmark by reformatting COCO~\cite{lin2014microsoft} and present performance of several baselines, which are variants of Panoptic FPN.
    \item We propose a novel framework for open-set panoptic segmentation, EOPSN, based on the exemplar theory, and demonstrate its effectiveness in detecting and segmenting examples in \unknown classes.
\end{itemize}

%% file: sections/2_related.tex
%!TEX root = ./../main.tex

\section{Related Works}\label{sec:related}

\subsection{Panoptic Segmentation}
Panoptic segmentation, a joint problem of semantic segmentation and instance segmentation, has received a lot of attention since Kirillov~\etal~\cite{kirillov2019panoptic} introduce the task.
There exist a large number of works for this problem and they are categorized into two groups: top-down and bottom-up approaches.
The top-down techniques~\cite{fu2019imp,kirillov2019panopticfpn,lazarow2020learning,li2019attention,liu2019end,porzi2019seamless,xiong2019upsnet} typically generate object proposals and segment the proposals before combining semantic segmentation results.
Mask R-CNN~\cite{he2017mask} is often deployed for instance segmentation and an encoder-decoder architecture is utilized for semantic segmentation.
On top of that, AUNet~\cite{li2019attention} leverages mask-level attention to transfer knowledge from the instance segmentation head to the semantic segmentation head.
UPSNet~\cite{xiong2019upsnet} proposes a parameter-free panoptic head to resolve conflicts in \thing and \stuff predictions.
On the other hand, the bottom-up~(proposal-free) methods obtain semantic segmentation outputs and then perform instance partitioning~\cite{cheng2020panoptic,wang2020axial,yang2019deeperlab}.
The bottom-up approaches are free from the instance-instance or instance-background overlap issues, but they generally achieve lower accuracy than the top-down methods.
Our method is categorized to a top-down approach, which processes instances and semantic backgrounds separately.

\subsection{Open-Set Learning}
Open-set recognition receives a spotlight in the computer vision community recent years.
The goal of open-set image classification is to classify \known class images observed during training while recognizing examples in \unknown classes~\cite{scheirer2012toward}.
In an open-set classification algorithm~\cite{scheirer2014probability}, \unknown class is further divided into two subgroups depending on whether it has been exposed as a negative class during training~(\seen-\unknown class) or it has never appeared~(\unseen-\unknown class).
On the other hand, similar to open-set object detection~\cite{dhamija2020overlooked}, such a strategy is not applicable to open-set panoptic segmentation since there is no information about whether unlabeled objects are seen or unseen in the training dataset.

Most open-set image classification methods~\cite{bendale2016towards,liu2019large,neal2018open,perera2020generative} make predictions to \unknown classes at test time if the probabilities of all \known classes are below a certain threshold.
OpenMax~\cite{bendale2016towards} leverages $(C+1)$-way classifier for $C$-class classification with background and trains the classifier using the Weibull distribution.
Generated images~\cite{perera2020generative} and counterfactual images~\cite{neal2018open} are employed to estimate robust decision boundaries for \known classes and identify \unknown classes effectively.
Liu~\etal~\cite{liu2019large} propose a memory- and clustering-based model for open-set image classification when the distribution of \known classes is long-tailed.

Recently, the open-set scenario is spread to other tasks such as object detection~\cite{dhamija2020overlooked,miller2018dropout} and semantic instance segmentation~\cite{pham2018bayesian}.
Miller~\etal~\cite{miller2018dropout} first address open-set object detection using dropout sampling.
The task is further investigated and formalized by Dhamija~\etal~\cite{dhamija2020overlooked}.
Pham~\etal~\cite{pham2018bayesian} introduce a Bayesian optimization framework that considers both object boundaries and masks for open-set instance segmentation.
This work formulates the open-set panoptic segmentation task on top of the closed-set counterpart.

\subsection{Exemplar-Based Learning}
Unlike prototype-based techniques~\cite{snell2017prototypical,xie2018learning} that utilize cluster centroids as the representatives of the individual clusters, exemplar-based methods directly employ samples stored in memory.
One of the well-known exemplar-based methods in machine learning is $k$-nearest neighbor algorithm~\cite{cover1967nearest}, which assigns labels of new data using the training examples in neighborhood.
Wu~\etal~\cite{wu2018unsupervised} perform representation learning without manual annotations by providing each example with a separate class label.
Incremental learning frameworks~\cite{guo2020exemplar,prabhu2020gdumb,rebuffi2017icarl} often adopt exemplar-based learning to prevent catastrophic forgetting~\cite{mccloskey1989catastrophic}.
They store a small number of exemplars that represent the whole dataset or classes in the previously tasks effectively.
MemAE~\cite{gong2019memorizing} employs exemplar-based learning for anomaly detection.

Existing exemplar-based learning approaches assume that ground-truth labels for exemplars are available or make the information less critical.
On the other hand, our problem is much more challenging because it requires to identify exemplars for diverse \unknown classes automatically and learn their representations properly without labels.

%% file: sections/3_ops.tex
%!TEX root = ./../main.tex
\section{Open-Set Panoptic Segmentation~(OPS)}\label{sec:open-set_panoptic_seg}
This section defines the open-set panoptic segmentation with its evaluation metrics.
Then, we discuss the inherent challenges of the task and how we make the problem tractable via reasonable assumptions. 

\subsection{Definition of Label and Task}
\label{subsec:def}

The open-set panoptic segmentation is similar to the closed-set method.
The key difference is the existence of \unknown classes, which are not available for training but appear in testing while the \known classes are always available.
Another criterion to distinguish class types is how an object is formed physically;
the \textit{thing} class, denoted by $\mathcal{C}^{\text{Th}}$, consists of the objects with concrete shape and structure while the \textit{stuff} class, $\mathcal{C}^{\text{St}}$, involves amorphous background regions (\eg, \cls{sky} or \cls{sand}) or unstructured objects~(\eg, \cls{tree} or \cls{grass}).
In addition, we introduce another kind of semantic label, called \void, which is not annotated in training data and corresponds to ambiguous or out-of-class pixels.

Given a predefined set of $C$ semantic classes encoded by $\mathcal{C} :=  \{0, ... , C-1\}$ and a set of \unknown class codes denoted by $\mathcal{U}$, the $i^\text{th}$ pixel of an image is labeled by a tuple, $(l_i, z_i) \in (\mathcal{C} \cup \ \mathcal{U}) \times \mathbb{N}$, where $l_i$ and $z_i$ indicate a semantic class label and its instance identifier, respectively.
In our problem definition, $\mathcal{U}$ actually has a single element that represents the entire \unknown classes.
The group of pixels with the same instance identifier ($z_i$) constitute a segment that belongs to class $l_i \in (\mathcal{C} \cup \ \mathcal{U})$.
In summary, the open-set panoptic segmentation aims to find all segments with \known labels or \unknown flags in a given image.

%%%%%
%
\begin{figure}[t]
\begin{center}
\includegraphics[width=1.0\linewidth] {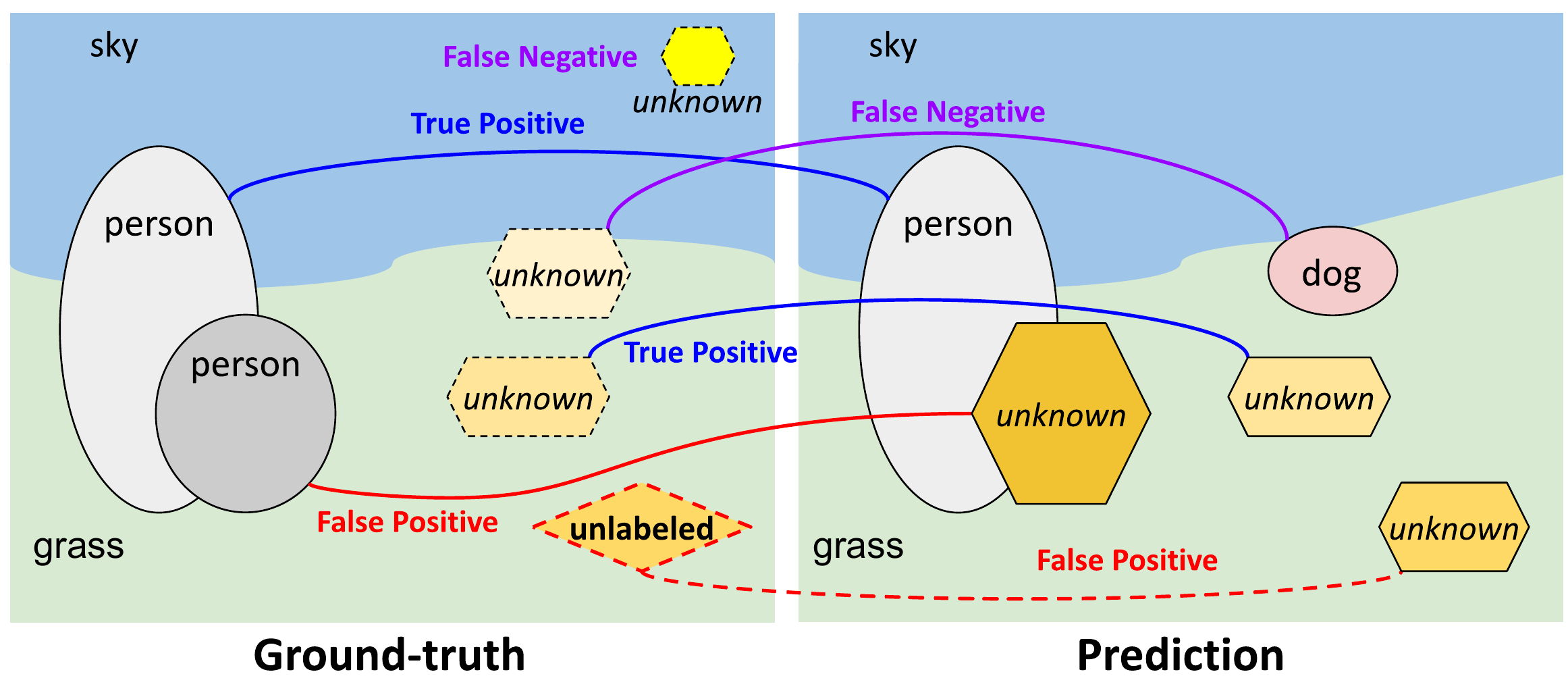}
\end{center}
\vspace{-0.5cm}
   \caption{
   Toy example illustration of ground-truth and predicted panoptic segmentation of an image.
   Objects with black-dotted border denote \unknown classes and an object with red-dotted border in the ground-truth is an unlabeled object.
The connections across objects in the two different images indicate matching results with their  attributes.
 Although the model correctly finds a rightmost \unknown instance, it is considered as a false positive since the instance is not labeled in the ground-truth.}
   
\label{fig:toy}
\end{figure}
%
%%%%%

\begin{figure*}[t]
\begin{center}
\includegraphics[width=1\linewidth] {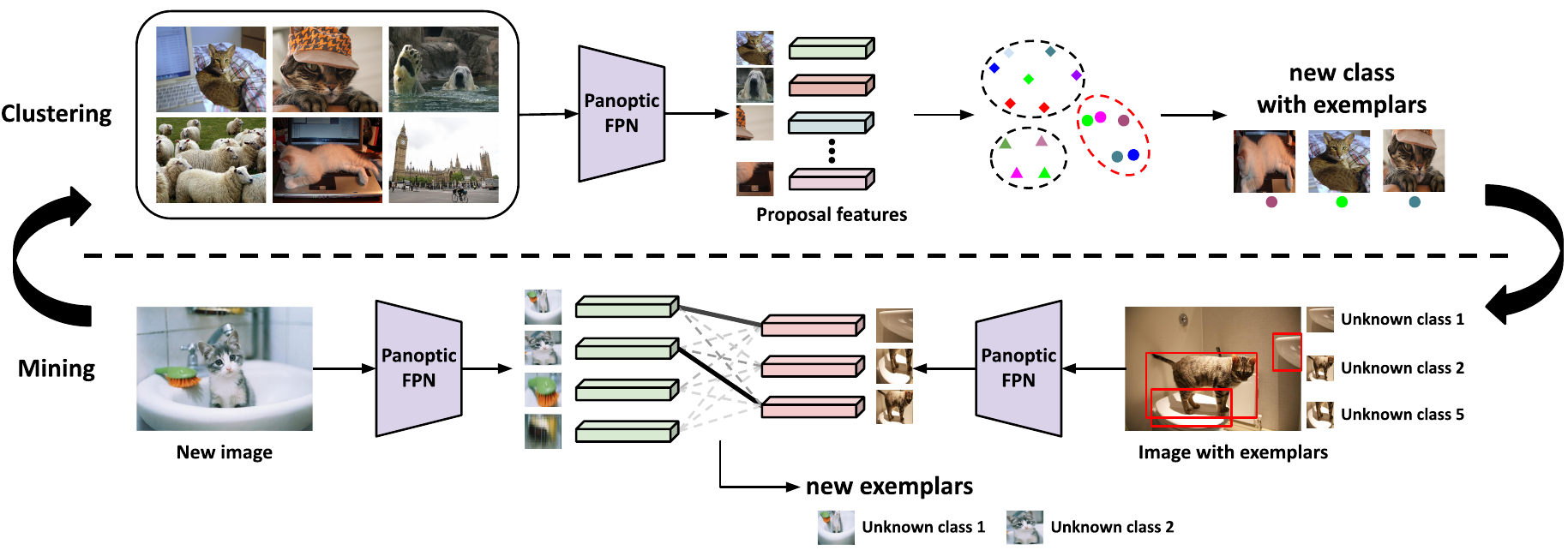}
\end{center}
\vspace{-0.4cm}
   \caption{
   Illustration of the proposed framework, EOPSN.
The model first identifies \unknown classes with the corresponding exemplars using $k$-means clustering~\cite{macqueen1967some} of the detected bounding boxes in \void areas from a small subset of images~(\textit{clustering} stage).
Each element in a cluster denotes a proposal feature while each color indicates a source image.
The cluster in red-dotted ellipse corresponds to one of the new \unknown classes and each image patch in the cluster becomes an exemplar.
Identified \unknown classes and their exemplars are used to discover more exemplars by comparing with object proposals of an input image in subsequent iterations~(\textit{mining} stage).
The two stages alternate to identify and augment \unknown classes but clustering is performed sparsely to reduce computational complexity.
    }
\label{fig:model}
\end{figure*}

\subsection{Evaluation Metric}
We utilize standard panoptic segmentation metrics~\cite{kirillov2019panoptic}, which include panoptic quality~(PQ), segmentation quality~(SQ) and recognition quality~(RQ) for both \known and \unknown classes.
The metrics are defined as
\begin{equation}
\small{\text{PQ}} = \underbrace{\frac{\sum_{(p, g) \in \TP} \text{IoU}(p, g)}{\vphantom{\frac{1}{2}}|\TP|}}_{\text{segmentation quality (SQ)}} \cdot \underbrace{\frac{|\TP|}{|\TP| + \frac{1}{2} |\FP| + \frac{1}{2} |\FN|}}_{\text{recognition quality (RQ)}},\\
\end{equation}
where IoU means Intersection-over-Union of two regions, and \textit{TP}, \textit{FP}, and \textit{FN} denote true positive, false positive, and false negative, respectively, as illustrated in Figure~\ref{fig:toy}.
If a predicted segment and a ground-truth region with the same semantic label are highly overlapped, \ie IoU $>0.5$, the identified segment is classified as the true positive set, \textit{TP}.

\subsection{Challenges}\label{subsec:challenges}

\paragraph{Finding unknown instances}
Unlike the open-set image classification, which simply recognizes unknown class images, an open-set panoptic segmentation model needs to find objects in \unknown class from each image.
It is a very difficult problem since there is no explicit semantic knowledge about the objects and the \unknown objects might have been labeled as background during training.

\vspace{-0.25cm}
\paragraph{Ground-truth labeling} 
Panoptic segmentation requires pixel-level annotations for each instance and semantic background region.
Annotators are supposed to delineate all instances and background stuffs thoroughly.
However, unfortunately, a certain object may consist of multiple components that correspond to other object classes, making comprehensive annotations challenging.
For example, \cls{car} is composed of several parts such as windows, side mirrors, lights, tires, etc.
Also, the regions in some \stuff labels (\eg, \cls{tree}, \cls{gravel}) can be divided into several instances.
Without a concrete guideline, annotations will be inconsistent and unreliable.
It hinders training open-set panoptic segmentation models and measuring their performance.

\vspace{-0.25cm}
\paragraph{Evaluation}
Current metrics for evaluating panoptic segmentation performance assume that each image has a complete annotation.
However, the universe set of class labels is unbounded in practice, which makes the evaluation inconsistent with true quality of panoptic segmentation.
In other words, if an open-set panoptic segmentation algorithm identifies \unknown instances very accurately but the labels of the detected objects are missing as illustrated in Figure~\ref{fig:toy}, the false positive rate will be large, which leads to low RQ and PQ values.
On the other hand, SQ does not suffer from such an issue but it is not straightforward to represent the overall performance since it only considers true positives.

\subsection{Tractable Problem Definition}\label{sec:problem_def}

Since we face significant conceptual and practical challenges in OPS as discussed in Section~\ref{subsec:challenges}, we propose a tractable version of problem definition as follows.
\vspace{-0.25cm}
\paragraph{Assumption}
We have the following three assumptions regarding \unknown classes observed in training data.
First, all \unknown classes belong to the \thing category.
This is because potential \unknown classes in the \stuff category are often ill-defined.
The second assumption is that parts of \known classes cannot be \unknown classes.
For example, assuming that \cls{car} and \cls{person} are \known, \cls{tire} and \cls{head} cannot be \unknown instances since they are parts of the aforementioned \known classes, respectively.
However, if \cls{tire} exists by itself, not as a part of another, it can be an \unknown class.
The final assumption is that \unknown class objects only appear in the \void regions during training.
The purpose of the assumptions is to prevent confusion between \known and \unknown class regions.

\vspace{-0.25cm}
\paragraph{Dataset}
Perfect annotation for the open-set panoptic segmentation is not realistic.
Hence, a reasonable way to measure the OPS quality is training on one dataset and testing on another with \unknown classes, which is similar to multi-domain semantic segmentation~\cite{lambert2020mseg}.
However, there is a domain gap between datasets and the definition of an \emph{object} may be different in each dataset.
Therefore, we create a dataset for OPS using an existing dataset.
There are several public benchmarks for panoptic segmentation, which include COCO~\cite{lin2014microsoft}, Cityscape~\cite{cordts2016cityscapes}, Mapillary~\cite{neuhold2017mapillary} and ADE20k~\cite{zhou2017scene}.
However, all of them except COCO are specialized to specific environments such as driving or indoor scene.
Hence, we adopt COCO, which contains natural photographs in everyday life.
We generate several splits with different numbers of \unknown classes sampled from the  \thing classes.
However, our benchmark cannot evaluate unlabeled \unknown classes properly as discussed earlier because the label information is completely missing throughout training and evaluation.

%% file: sections/4_method.tex
%!TEX root = ./../main.tex

\section{Method}\label{sec:method}
We explain the baseline models and the proposed approach for open-set panoptic segmentation in this section.

\subsection{Motivation and Overview}
The core technique in OPS is how to find instances in \unknown classes.
One possible way is to leverage a class-agnostic model such as RPN~\cite{ren2015faster}, which predicts an objectness score.
We employ this approach as our baseline.
However, it has a critical limitation; the score is optimized to the \known instances. 
Hence, the performance of the model depends heavily on the semantic similarity between \known and \unknown classes, which means that \unknown classes are unlikely to be recognized by the model if they are not semantically related to any of \known classes.

To tackle this drawback, we propose an exemplar-based learning framework on top of the baseline to find \unknown classes from training data more effectively.
Note that an exemplar means a bounding box corresponding to an example identified as \unknown class.
Our model first finds \unknown classes with the associated exemplars using $k$-means clustering~\cite{macqueen1967some}, which is applied to the detected bounding boxes within \void area.
A tightly coupled cluster with a high objectness score is considered as an \unknown class, and each image patch in the cluster becomes an exemplar.
The exemplars are used to find new ones by comparing the similarities between the existing exemplars and the object proposals from the images contained in the subsequent mini-batches.
They are also employed as the pseudo-ground-truth bounding boxes for the future iterations.
The clustering and mining procedures alternate to detect new \unknown classes and collect examples with the identified \unknown labels while the backbone network is concurrently optimized with both types of class labels.
Figure~\ref{fig:model} illustrates the procedure of the proposed approach.

\subsection{Baseline}\label{subsec:baseline}
Since only \thing classes can be \unknown according to our problem definition, it is reasonable to adopt top-down panoptic segmentation methods with RPN as baseline models.
We choose Panoptic FPN~\cite{kirillov2019panopticfpn}, which is composed of an instance segmentation head~\cite{he2017mask} and a semantic segmentation head on top of a shared feature extractor.
The instance segmentation head is exactly the same as Mask R-CNN~\cite{he2017mask} and the semantic segmentation head is identical to the FPN-based decoder.

We modify the bounding box regressor and the mask predictor in the instance segmentation head to make them class-agnostic for handling \unknown classes.
This baseline model first predicts bounding boxes pertaining to \known classes and finds \unknown instances based on the objectness scores of the candidates from RPN, where the score threshold is $0.5$.
We also introduce the \void class in the classification branch within the instance segmentation head to identify the bounding boxes sampled from the \void regions.
For the supervision of the \void classes, we use the bounding boxes that more than a half of the region is inside the \void area. 
Note that, since the boxes from \void regions do not necessarily correspond to objects, they are not employed to train the RPN.

%%%%%%%%%%%%%%%%%%%%%%%%
%
\input{tables/baseline.tex}
%
%%%%%%%%%%%%%%%%%%%%%%%%

\subsection{Exemplar-Based Open-Set Learning}
The main goal of our exemplar-based learning is to identify bounding boxes with coherent features that belong to the same \unknown classes.
Our exemplar-based learning consists of two stages: clustering candidate proposals in \unknown classes obtained from a subset of training images, which aims to find new categories and their exemplars, and mining new exemplars via similarity matching with the existing ones.
These clustering and mining stages alternate throughout the training procedure.

For the clustering and mining stages, we extract the features from \unknown candidate bounding boxes in \void regions for \unknown classes.
To this end, we first reduce duplicate detections by applying Non-Maximum Suppression~(NMS) with the IoU threshold $1\times10^{-7}$, and then sample candidate proposals with the weights based on the objectness scores given by RPN~\cite{ren2015faster}.
After that, 1024-dimensional features are obtained from the proposals using a backbone network, and we perform a series of operations including RoI-Align, GAP, and feature computation using two \textsf{fc} layers as in~\cite{he2017mask}.

\vspace{-0.25cm}
\paragraph{Clustering}
To find \unknown classes, we perform $k$-means clustering based on the consine distance using the extracted features from candidate bounding boxes.
Note that the clustering is performed at every 200 iterations using the features computed for all the examples presented in the last 200 mini-batches.
We generate a large number of clusters~(over-clustering) and take a subset of clusters only that clearly correspond to true \unknown classes.
Such a cluster should have a high average objectness score but a small average cosine distance between the centroid and all elements since the instances in a true \unknown class are prone to be clustered tightly while non-object proposals have loose connections and low objectness scores.
We identify exemplars from the high-quality clusters and store them in the subsequent mining stage.
Note that we maintain exemplars in multiple \unknown classes during training but collapse all of them and make a single \unknown class for evaluation.

\vspace{-0.25cm}
\paragraph{Mining exemplars}
Our approach mines additional exemplars with the \unknown concepts detected in the past, from the images in the incoming mini-batch.
This can be done easily by comparing features of the stored exemplars and the features of the object proposals from the images in a new mini-batch.
Note that, since the feature extraction network is updated over time, we need to recompute the features of the stored exemplars.
We accept the proposals generated by RPN as new exemplars if their cosine similarities to any of existing exemplars are higher than a threshold.

\vspace{-0.25cm}
\paragraph{Loss}
We utilize the almost same classification loss (including the regression loss), mask loss, and semantic segmentation loss adopted in Panoptic FPN~\cite{kirillov2019panopticfpn}.
The only difference is classification loss on the instance segmentation head.
We utilize the cross-entropy loss over \known classes, a background class~(bg), and \unknown classes as follows:
\begin{equation}
    \Lcal_\text{ce} =  \sum_{c \in (\mathcal{C}^{\text{Th}} \cup \{\text{bg} \}\cup \mathcal{U}) } -y_c \log p_c,
    \label{eq:cls}
\end{equation}
where $y_c$ is a (pseudo-)ground-truth label and $p_c$ is the softmax score of class $c$.
Additionally, we give negative supervision to the object proposals in the \void regions so that they are not classified as \known classes, which is given by
\begin{equation}
    \Lcal_\text{void} = \sum_{c \in \mathcal{C}^{\text{Th}}} -\log (1-p_c).
    \label{eq:neg}
\end{equation}
Then, the  total classification loss is given by
\begin{equation}
\mathcal{L}_\text{cls} = \mathcal{L}_\text{ce} + \mathbb{I}_\text{void} \mathcal{L}_\text{void},
\label{eq:total_loss}
\end{equation}
where $\mathbb{I}_\text{void}$ is an indicator function for a box in \void regions.

\vspace{-0.25cm}
\paragraph{Inference}
Our model predicts instance segmentation and semantic segmentation results and then combine them to make a panoptic segmentation output.
We first generate panoptic segmentation map using a same inference mechanism of the standard panoptic segmentation model~\cite{kirillov2019panopticfpn} and then add predicted \unknown instances additionally.

%% file: tables/baseline.tex
%!TEX root = ./../main.tex

\begin{table*}[t]
\caption{
Open-set panoptic segmentation results on the COCO \val set~($K = 20\%$) of the baseline approaches with different utilization of \void regions: 
used as backgrounds~(\textit{Void-background}), ignored~(\textit{Void-ignorance}), supervised by Eq~\eqref{eq:neg}~(\textit{Void-suppression}) and trained as a new label~(\textit{Void-train}) during training.
}
\label{tab:void}
\centering
\scalebox{0.9}{%
\setlength{\tabcolsep}{4pt}
\begin{tabular}{@{}c|ccc|ccc|ccc|ccc@{}}
\hline
\toprule
\multirow{2}{*}{Utilization of \void regions} & \multicolumn{9}{c|}{Known} & \multicolumn{3}{c}{Unknown} \\ \cline{2-13}
  & \addstackgap[3.5pt]{PQ} & SQ &RQ & $\text{PQ}^{\text{Th}}$ & $\text{SQ}^{\text{Th}}$& $\text{RQ}^{\text{Th}}$ & $\text{PQ}^{\text{St}}$ & $\text{SQ}^{\text{St}}$ & $\text{RQ}^{\text{St}}$ & PQ & SQ &RQ \\ \hline\hline
Void-background & \textbf{37.7} & 76.3 & \textbf{46.6} & 44.8 & 79.3  & 54.1 & \textbf{29.2} & 72.8  & \textbf{37.5} & 4.0 & 71.1 & 5.7 \\
Void-ignorance & 37.2 & 76.3 & 45.9 & 43.9 & 79.0 & 53.1 & 29.1 & \textbf{73.0} & 37.3 & 3.7 & 71.8 & 5.2 \\
Void-suppression & 37.5 & 75.9 & 46.1 & \textbf{45.1} & \textbf{80.6} & \textbf{54.5} & 28.2 & 70.2 & 36.1 & 7.2 & \textbf{75.3} & 9.6 \\
 Void-train & 36.9 & \textbf{76.4}  & 45.5 & 44.0 & 80.3  & 53.3 & 28.2 & 71.7  & 36.0 & \textbf{7.8} & 73.4 & \textbf{10.7} \\
\bottomrule
\end{tabular}
} 
\vspace{-0.2cm}
\end{table*}

%% file: sections/5_exp.tex
%!TEX root = ./../main.tex

%%%%%%%%%%%%%%%%%%%%%%%%%%%
%
\input{tables/main.tex} % tab:void
%
%%%%%%%%%%%%%%%%%%%%%%%%%%%

\section{Experiments}\label{sec:exp}
We describe our experimental setting and present the results.
We also analyze various aspects of the proposed framework.
Refer to the supplementary document for implementation details and more experimental results.

\subsection{Dataset and Evaluation}
All experiments are conducted on COCO~\cite{lin2014microsoft}.
We utilize 2017 panoptic segmentation train and validation splits, which contain 118K and 5K images, respectively, with annotations of 80 \thing classes and 53 \stuff classes.
To construct an open-set setup, we remove annotations for a subset of \known thing classes in the training dataset and consider them as \unknown classes.
We construct three different splits with varying numbers of \unknown classes (5\%, 10\%, 20\%)\footnote{We sorted the class labels based on their frequencies and sampled a subset of labels regularly for removal to simulate \unknown classes.}.
The removed classes in the the splits are shown below, where the classes are removed cumulatively.
\begin{itemize}
	\item 5\%: \cls{car}, \cls{cow}, \cls{pizza},  \cls{toilet} \vspace{-0.2cm}
	\item 10\%: \cls{boat}, \cls{tie}, \cls{zebra},  \cls{stop sign} \vspace{-0.2cm}
	\item 20\%: \cls{dining table}, \cls{banana}, \cls{bicycle},  \cls{cake}, \cls{sink}, \cls{cat}, \cls{keyboard}, \cls{bear}
\end{itemize}

We employ the standard panoptic segmentation metrics, \ie, PQ, SQ, and RQ, to evaluate performance.
We report performance of \known class and \unknown class separately.

\subsection{Quantitative Results}

Table~\ref{tab:void} presents open-set panoptic segmentation performance of the baseline models relying on Panoptic FPN by varying the utilization of \void regions on COCO \val set with 20\% \unknown classes.
There are four different usages of the boxes: training as backgrounds~(\textit{Void-background}), ignoring the boxes~(\textit{Void-ignorance}), providing the negative supervision for \known classes  using \eqref{eq:neg} to prevent the boxes from being classified as \known class~(\textit{Void-suppression}), and training as a new class with standard cross-entropy loss~(\textit{Void-train}).
All variations have similar performance for \known classes but \unknown classes have different characteristics.
\textit{Void-ignorance} model has the worst performance since the instances in \unknown class are classified as \known classes.
\textit{Void-suppression} and \textit{Void-train} models achieve comparable accuracy and outperform the rest two options  since bounding boxes in \void area are not trained as backgrounds in RPN and do not have low objectness scores.

Table~\ref{tab:main} summarizes the experimental results on the COCO \val set with different \known-\unknown splits.
The supervised model denotes Panoptic FPN trained on all classes without the \unknown ones.
The baseline is Panoptic FPN with the \textit{Void-}\train option, which is one of the variants presented in Table~\ref{tab:void}.
EOPSN outperforms the baseline method in all aspects for \unknown classes with large margins while still achieving competitive performance for \known classes.
Overall, both PQ and RQ  in \unknown classes are much smaller than \known cases while SQ's  are similar in both \known  and \unknown classes. 
This is because SQ is computed based only on true positives.

\subsection{Qualitative Results}
\begin{figure}[t]
\begin{center}
\includegraphics[width=1\linewidth] {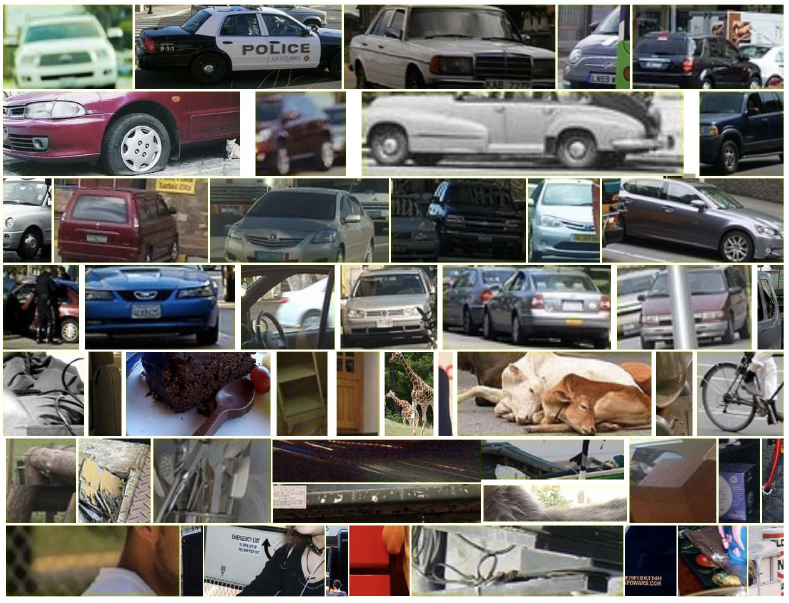}
\end{center}
\vspace{-0.3cm}
   \caption{
Visualization of the exemplars in an identified \unknown class by the first clustering in EOPSN.
   }
\label{fig:exemplar}
\end{figure}

Figure~\ref{fig:exemplar} illustrates the exemplars in a detected \unknown class after the first clustering with $K=20$\% in EOPSN.
Most exemplars contain \cls{car} while there exist objects in other \unknown classes such as \cls{cake}, \cls{cow}, and \cls{bicycle} as well as instances in \known class, \cls{giraffe}.

\begin{figure*}[t]
\begin{center}
\includegraphics[width=0.95\linewidth] {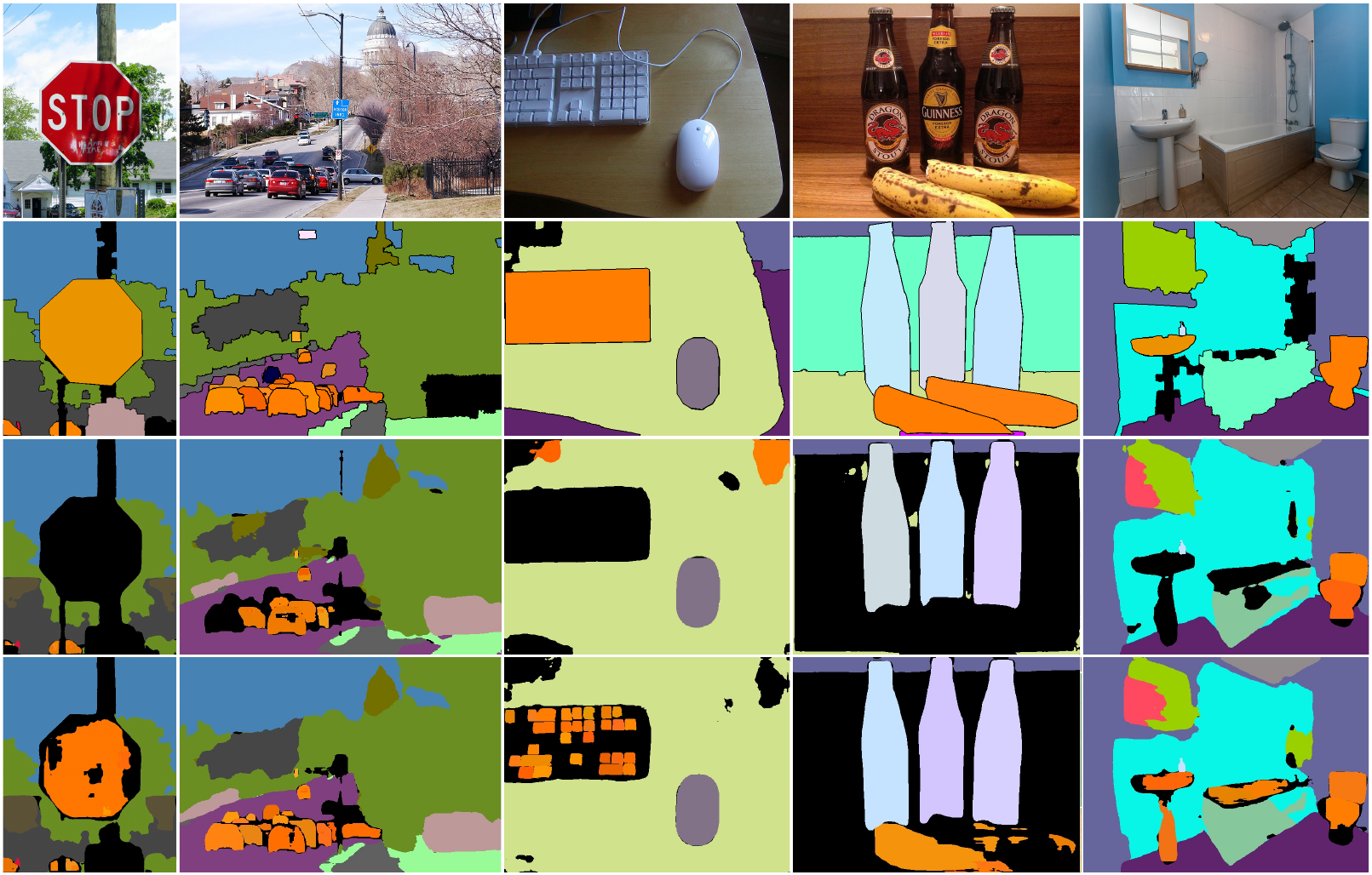}
\end{center}
\vspace{-0.3cm}
   \caption{
   Qualitative results on the COCO \val set with $K=20\%$. The first row presents input images and the subsequent rows illustrate ground-truths, results of the baseline (Void-train), and those of EOPSN.
Instances in the \unknown class are denoted by orange color.
   }
\label{fig:qual}
\end{figure*}

Figure~\ref{fig:qual} presents the comparison of the open-set panoptic segmentation methods. between the baseline model~(the third row) and EOPSN~(the forth row) on the COCO \val set with $K=20$\%. 
The \unknown classes in the figures are \cls{stop sign}, \cls{car}, \cls{keyboard}, \cls{sink}, and \cls{toilet}.
The second row shows ground-truths, where \unknown classes are in orange.
We observe that EOPSN successively finds several \unknown instances missed by the baseline model. 
Interestingly, EOPSN discovers the keys inside the keyboard and the bathtub unit, which are not included in the COCO classes.

%% file: tables/main.tex
%!TEX root = ./../main.tex

\begin{table*}[t]
\centering
\caption{Open-set panoptic segmentation results on the COCO \val set with several different \known-\unknown splits $K$ denotes the ratio of \unknown classes to all classes.
The numbers in bold denote higher scores than the opponents.
%\vspace{-0.2cm}
}
\label{tab:main}
\scalebox{0.9}{%
\setlength{\tabcolsep}{4pt}
\begin{tabular}{@{}cc|ccc|ccc|ccc|ccc@{}}
\hline
\toprule
\multirow{2}{*}{$K(\%)$} & \multirow{2}{*}{Model} &  \multicolumn{9}{c|}{Known} & \multicolumn{3}{c}{Unknown} \\ \cline{3-14}
&  & \addstackgap[3.5pt]{PQ} & SQ & RQ & $\text{PQ}^{\text{Th}}$ & $\text{SQ}^{\text{Th}}$& $\text{RQ}^{\text{Th}}$ & $\text{PQ}^{\text{St}}$ & $\text{SQ}^{\text{St}}$ & $\text{RQ}^{\text{St}}$ & PQ & SQ &RQ \\ \hline\hline
 & Supervised & 39.4 & 77.7 & 48.4 & 45.8 & 80.7 & 55.4 & 29.7 & 73.1 & 38.0 &- &- &- \\ \hline
%\multirow{2}{*}{5} & Baseline & 37.5 & 77.5 & 46.1 & 44.0 & 80.5 & 53.2  & 28.2 & \textbf{73.2} & 35.9& 5.3 & 74.0 & 7.2  \\
\multirow{2}{*}{5} & Baseline~(\textit{Void-train}) & 37.7 & 76.7 & 46.4 & 44.2 & 80.4 & 53.5  & 28.3 & 71.3 & 36.2& 10.0 & 73.8 & 13.5  \\
& EOPSN & \textbf{38.0} & \textbf{76.9} & \textbf{46.8} & \textbf{44.8} & \textbf{80.5} & \textbf{54.2} & 28.3 & \textbf{71.9} & 36.2 & \textbf{23.1} &\textbf{ 74.7} & \textbf{30.9} \\ \hline
%& EOPSN & 38.1 & \textbf{76.8} & 46.8 & 45.0 & 80.4 & \textbf{54.4} & 28.1 & 71.6 & 36.0 & \textbf{16.4} & \textbf{77.4} & \textbf{21.1} \\ \hline
%$\multirow{2}{*}{10}  & Baseline &  37.1 & 76.6 & 45.7 & 43.7 & 80.1 & 53.0 & 28.1 & 71.7 & 35.8 & 4.8 & 72.5 & 6.6\\
\multirow{2}{*}{10}  & Baseline~(\textit{Void-train}) &  36.9 & 75.4 & 45.5 & 43.2 & 79.0 & 52.4 & 28.3 & 70.4 & 36.2 & 8.5 & 73.2 & 11.6 \\
& EOPSN &  \textbf{37.7} & \textbf{76.8} & \textbf{46.3} & \textbf{44.5} & \textbf{80.6} & \textbf{53.8} & \textbf{28.4} & \textbf{71.8} & 36.2 & \textbf{17.9} & \textbf{76.8} & \textbf{23.3}\\ \hline
%& EOPSN &  37.7 & 76.8 & 46.4 & 44.1 & 80.4 & 53.9 & 28.3 & 71.9 & 36.2 & \textbf{11.1} & \textbf{74.6} & \textbf{14.9}\\ \hline
%\multirow{2}{*}{20} & Baseline & 36.5 & 76.7  & 45.2 & 43.7 & 80.0  & 53.0 & 27.9 & \textbf{72.6}  & 35.6 & \weird{7.2} & \weird{73.1} & \weird{14.7} \\
\multirow{2}{*}{20} & Baseline~(\textit{Void-train}) & 36.9 & \textbf{76.4}  & 45.5 & 44.0 & 80.3  & 53.3 & 28.2 & \textbf{71.7}  & 36.0 & 7.8 & 73.4 & 10.7 \\
& EOPSN & \textbf{37.4} & 76.2 & \textbf{46.2} & \textbf{45.0} & 80.3 & \textbf{54.5} & 28.2 & 71.2 & \textbf{36.2} & \textbf{11.3} & \textbf{73.8} & \textbf{15.3} \\
%& EOPSN & 37.5 & 76.3 & \textbf{46.3} & \textbf{45.4} & 80.3 & \textbf{54.9} & 28.1 & 71.5 & 35.9 & \textbf{8.8} & \textbf{75.1} & 11.7 \\
\bottomrule
\end{tabular}
}
\vspace{0.1cm}
\end{table*}

%% file: sections/6_discussion.tex
%!TEX root = ./../main.tex

\section{Discussion}\label{sec:discussion}

In addition to the challenges discussed in Section~\ref{subsec:challenges}, several critical issues still remain.
First of all, EOPSN mainly focuses on the classification branch in the instance segmentation head while we have not explored potential of the semantic segmentation head sufficiently.
We believe that combining a bottom-up approach to form \unknown segments would improve performance.

Second, EOPSN is based on clustering with training data to find new \unknown classes.
If an \unknown class does not appear in the training dataset, its performance will be degraded considerably.
However, as we collect more data, the number of \seen-\unknown classes will grow rapidly and the benefit of the proposed model will become salient; the techniques based on the class-agnostic objectness scores, \ie, the baseline methods based on RPN, would have less merits since they only employ the ground-truth information of \known classes.

Finally, a new metric should be defined for OPS.
If a trained model finds more \unknown instances in testing, which are legitimate but unlabeled, the inconsistency between true and measured accuracy will be aggravated.
A straightforward solution is to employ human evaluation, but it is expensive and may incur another critical issue related to consistency.
This issue may be alleviated by introducing a new metric by revising SQ.
For example, by properly considering false positives of \unknown classes, we can represent overall segmentation quality more accurately.

%% file: sections/7_conclusion.tex
%!TEX root = ./../main.tex

\section{Conclusion}\label{sec:conclusion}
We introduced a novel task referred to as open-set panoptic segmentation (OPS), which involves \unknown classes that appear in testing while not considered during training.
The goal of this task is to obtain a correct panoptic segmentation map for the union of \known and \unknown classes.
We provided a new benchmark dataset on top of COCO~\cite{lin2014microsoft}.

To tackle the challenging task, we presented EOPSN, an exemplar-based open-set panoptic segmentation network as a solid baseline model.
Our framework first identifies new classes with the associated exemplars by clustering the proposals sampled from a small subset of images, and then discovers new exemplars progressively using the rest of the images during training.
EOPSN outperforms the baselines based on variants of Panoptic FPN~\cite{kirillov2019panopticfpn}.
Since EOPSN is a generic approach that can be incorporated into any top-down panoptic segmentation model, the integration of the state-of-the-art networks would lead to better performance.

We hope that this work draws the attention of the computer vision community to open-set problems beyond simple recognition tasks.
OPS would facilitate large-scale dataset collection that requires dense labeling and\ allow us to tackle more realistic tasks in challenging scenarios.

%% file: sections/8_supp.tex
%!TEX root = ./../1361.tex

\setcounter{section}{0}
\renewcommand{\thesection}{\Alph{section}}
\setcounter{table}{0}
\renewcommand{\thetable}{\Alph{table}}
\setcounter{figure}{0}
\renewcommand{\thefigure}{\Alph{figure}}

\section*{Appendix}

\section{Details}
\subsection{Implementation}

The implementation is based on Panoptic FPN~\cite{kirillov2019panopticfpn}.
The backbone is ResNet-50-FPN~\cite{he2016deep,lin2017feature}.
We utilize Pytorch~\cite{paszke2019pytorch} distributed and Detectron2~\cite{wu2019detectron2}.

\subsection{Training}\label{subsec:training}

For baseline models, we employ ImageNet pre-trained ResNet-50-FPN~\cite{he2016deep,lin2017feature}.
The initial learning rate is 0.04 with a linear warm-up~\cite{goyal2017accurate} and reduced by $0.1$ at 30K and 40K step.
The total training step is 45K~(namely $1\times$ schedule).
The weight decay is 0.0001 and momentum is 0.9.
Semantic segmentation head trains \void label as a new class.
EOPSN is fine-tuned from the baseline with Eq.~(2)~(\textit{Void-suppression}) maintaining the learning rate $0.0004$ for 30K steps.
All other hyper-parameters follow those in Detectron2~\cite{wu2019detectron2}.
All models are trained in 8 Titan V100 GPUs using synchronized SGD, with a mini-batch size of 4 images per GPU.

For exemplar-based learning, we utilize middle or large sized bounding boxes (\ie, the area is larger than $32^2$) to reduce noise.
We sample at most 20 object proposals in every mini-batch and generate $128$ clusters in every $200$ steps.
For finding high-quality clusters, only top $10\%$ clusters in terms of the average cosine similarity between the centroids and their elements are used.
The objectness score threshold for selecting high-quality clusters is starting from $0.9$ and it is linearly increased to $0.99$ depending on the number of found \unknown classes.
Cosine distance threshold for finding coupled elements in a cluster and mining new exemplars is $0.15$, $0.025$, respectively and it is slightly decreased to 0.01.

\section{Hyper-parameters}
We test the proposed model with various hyper-parameters: the number of clusters, clustering interval, bounding box size.
All experiments in this section are conducted with $K=10\%$.
While comparing one hyper-parameter, all other parameters are fixed as described in Section~\ref{subsec:training}.

\paragraph{The number of clusters}
Table~\ref{tab:supp_k} presents that the number of clusters affects the performance.
This might be due to that the number of noises in each cluster is increased in the small number of clusters setting and the intra-class variability in each cluster is decreased in the large number of clusters setting.
However, all are still higher than the baseline ($8.5, 73.2, 11.6$ for PQ, SQ, and RQ, respectively).

\paragraph{Clustering Interval}
Table~\ref{tab:supp_interval} shows that short clustering interval decreases RQ and PQ.
This is because the number of object proposals and the number of used images for clustering are decreased, which leads the clustering stage to be more vulnerable to the noise and be hard to find \unknown classes with exemplars correctly.

\begin{table}[t]
\centering
\caption{Sensitivity analysis about the number of clusters on COCO \val set where $K$ is $10\%$.}
\label{tab:supp_k}
 \resizebox{1\linewidth}{!}{
\setlength{\tabcolsep}{4pt}
\begin{tabular}{@{}c|ccc|ccc@{}}
\hline
\toprule
 \multirow{2}{*}{The number of clusters} &  \multicolumn{3}{c|}{Known} & \multicolumn{3}{c}{Unknown} \\ \cline{2-7}
 & PQ & SQ & RQ & PQ & SQ &RQ \\ \hline\hline
64 & 37.3 & 76.2 & 45.8 & 12.9 & 76.5 & 16.8 \\
128 (ours) & 37.7 & 76.8 & 46.3 & \textbf{17.9} & 76.8 & \textbf{23.3} \\ 
256 & 37.2 & 76.8 & 45.7 & 12.6 & \textbf{78.8} & 16.0 \\ 
\bottomrule
\end{tabular}
}
\end{table}

\begin{table}[t]
\centering
\caption{Sensitivity analysis about clustering intervals on COCO \val set where $K$ is $10\%$.}
\label{tab:supp_interval}
\resizebox{1\linewidth}{!}
{%
\setlength{\tabcolsep}{4pt}
\begin{tabular}{@{}c|ccc|ccc@{}}
\hline
\toprule
 \multirow{2}{*}{Clustering interval} &  \multicolumn{3}{c|}{Known} & \multicolumn{3}{c}{Unknown} \\ \cline{2-7}
 & PQ & SQ & RQ & PQ & SQ &RQ \\ \hline\hline
100     & 37.6 & 77.2 & 46.3 & 8.2 & \textbf{77.5} & 10.6 \\
200 (ours) & 37.7 & 76.8 & 46.3 & \textbf{17.9} & 76.8 & \textbf{23.3} \\ 
400         & 37.7 & 77.5 & 46.3 & 14.6 & 76.4 & 19.1 \\
\bottomrule
\end{tabular}
}
\end{table}

\paragraph{Size of Object Proposals}

Table~\ref{tab:supp_size} presents that using large and medium proposals achieves the best performance on COCO \val set where $K$ is $10\%$.
On the other hand, using small sized object proposals dramatically degrades recognition performance since it increases noise during clustering, which hinders finding correct \unknown class.

\begin{table}[t]
\centering
\caption{Effectivity of size of proposals on COCO \val set where $K$ is $10\%$.}
\label{tab:supp_size}
 \resizebox{1\linewidth}{!}{
\setlength{\tabcolsep}{4pt}
\begin{tabular}{@{}c|ccc|ccc@{}}
\hline
\toprule
 \multirow{2}{*}{Proposal size} &  \multicolumn{3}{c|}{Known} & \multicolumn{3}{c}{Unknown} \\ \cline{2-7}
 & PQ & SQ & RQ & PQ & SQ &RQ \\ \hline\hline
 Large & 37.7 & 77.5 & 46.4 & 13.5 & \textbf{78.1} & 17.3 \\
 Medium & 37.7 & 77.5 & 46.4 & 12.5 & 74.8 & 16.7 \\
 Small & 37.6 & 76.7 & 46.3 & 0.3 & 64.1 & 0.4 \\
 Large + Medium  & 37.7 & 76.8 & 46.3 & \textbf{17.9} & 76.8 & \textbf{23.3} \\ 
Large + Medium + Small & 37.8 & 77.1 & 46.6 & 6.9 & 69.8 & 9.9  \\
\bottomrule
\end{tabular}
}
\end{table}

\section{Qualitative Results}

\begin{figure*}[t]
\begin{center}
\includegraphics[width=0.92\linewidth] {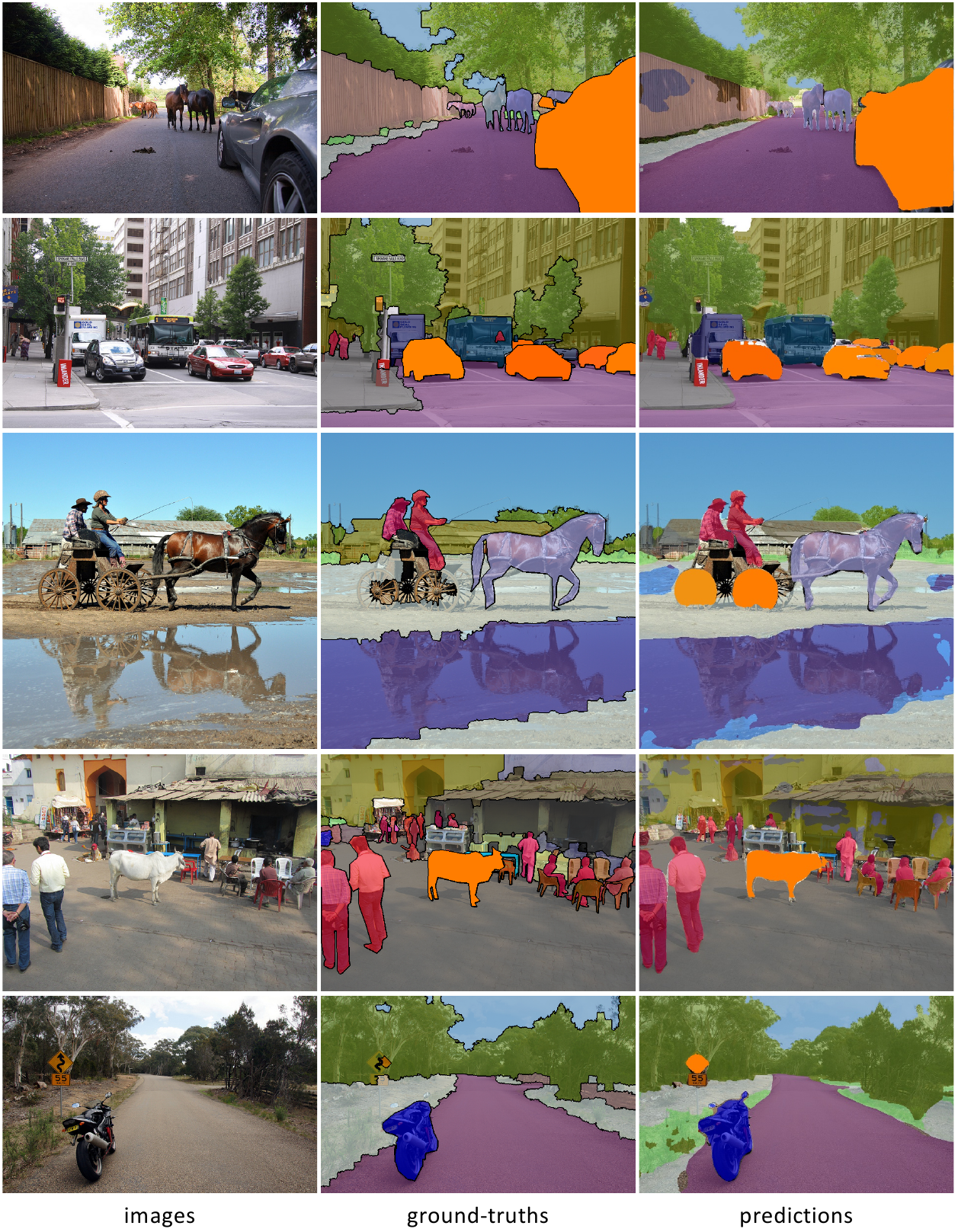}
\end{center}
   \caption{
   Qualitative results on COCO \val set with $K=10\%$. Each column denotes images, ground-truths and predictions of EOPSN, respectively.
Instances in the \unknown class are denoted by orange color.
   }
\label{fig:supp_qual}
\end{figure*}

Figure~\ref{fig:supp_qual} shows open-set panoptic segmentation results from EOPSN on COCO \val set with $K=10\%$.
Instances in the \unknown class are denoted by orange color.

